\documentclass[runningheads]{llncs}
\usepackage[T1]{fontenc}
\usepackage{graphicx}
\usepackage{marvosym}
\usepackage{multirow}
\usepackage{amssymb}
\usepackage{amsmath}
\begin{document}
\title{Neural-Symbolic Recommendation with Graph-Enhanced Information}
%
%
\author{Bang Chen\inst{1} \and
Wei Peng\inst{2} \and
Maonian Wu\inst{1}\textsuperscript{\Letter} \and
Bo Zheng\inst{1} \and
Shaojun Zhu\inst{1}}
\authorrunning{B. Chen et al.}
%
\institute{School of Information Engineering, Huzhou University, Huzhou, China \\
\email{cb\_cnzjhz@outlook.com, wmn@zjhu.edu.cn} \and
College of Computer Science, Guizhou University, Guiyang, China}
\maketitle              
\begin{abstract}
The recommendation system is not only a problem of inductive statistics from data but also a cognitive task that requires reasoning ability. The most advanced graph neural networks have been widely used in recommendation systems because they can capture implicit structured information from graph-structured data. However, like most neural network algorithms, they only learn matching patterns from a perception perspective. Some researchers use user behavior for logic reasoning to achieve recommendation prediction from the perspective of cognitive reasoning, but this kind of reasoning is a local one and ignores implicit information on a global scale. In this work, we combine the advantages of graph neural networks and propositional logic operations to construct a neuro-symbolic recommendation model with both global implicit reasoning ability and local explicit logic reasoning ability. We first build an item-item graph based on the principle of adjacent interaction and use graph neural networks to capture implicit information in global data. Then we transform user behavior into propositional logic expressions to achieve recommendations from the perspective of cognitive reasoning. Extensive experiments on five public datasets show that our proposed model outperforms several state-of-the-art methods, source code is avaliable at [https://github.com/hanzo2020/GNNLR].

\end{abstract}
\section{Introducation}
The explosive growth in internet information has made recommendation systems increasingly valuable as auxiliary decision-making tools for online users in various areas, including e-commerce~\cite{wang2020time}, video~\cite{liu2019user}, and social networks~\cite{liao2022sociallgn}. Classic recommendation methods mainly include matrix factorization-based approaches~\cite{rendle2009bpr}, neural network methods~\cite{he2017neural}, time-series-based methods~\cite{li2017neural}, and others that leverage richer external heterogeneous information sources, such as sentiment space context~\cite{zhou2022point} and knowledge graphs~\cite{yang2022knowledge}. Graph neural networks have recently gained attention for their success in structured knowledge tasks. They have been widely used in recommendation systems,  including Wang et al.'s graph collaborative filtering approach~\cite{wang2019neural}, He et al.'s lightweight graph collaborative filtering method~\cite{he2020lightgcn}, and Wang's recommendation algorithm based on a graph attention model~\cite{wang2019kgat}. The advantage of graph neural networks is that they can aggregate information from neighbor nodes through the data structure of graphs in a global view, which allows them to better capture implicit high-order information compared to other types of neural network methods.

Although the above methods have their advantages, they all have one obvious drawback: they only learn matching patterns in the data from a perception perspective, not reasoning~\cite{shi2020neural}. Although graph neural networks can utilize structured knowledge from graphs, such aggregation learning is essentially a weak reasoning mode within the scope of perceptual learning and does not consider explicit logic reasoning relationships between entities~\cite{chen2022graph}. As a task that requires logic reasoning ability, recommendation problems are more like decision-making processes based on past known information. For example, a user who has recently purchased a computer does not need recommendations for similar products but needs peripheral products such as keyboards and mice. However, in current recommendation system applications, users are often recommended similar products immediately after purchasing an item, even if their demand has already disappeared.

Some researchers have attempted to incorporate logic reasoning into recommendation algorithms to address the above issue. For example, Shi et al.~\cite{shi2020neural} proposed a neural logic reasoning algorithm that uses propositional logic to achieve recommendations. Subsequently, Chen et al.~\cite{chen2021neural} proposed neural collaborative reasoning and added user information to improve the model's personalized reasoning ability for users. However, these advanced methods only perform logic reasoning based on the current user's historical interaction behavior, which is just a local range of reasoning and lacks implicit high-order information from the global perspective. Especially when there is an enormous amount of recommendation data available, the number of items interacted with by a single user compared to all items is usually very small; therefore, it is evident that large amounts of implicit global information will be ignored.

To address the above challenges, this paper proposes a neural-symbolic recommendation model based on graph neural networks and Proposition Logic. We use logic modules to compensate for the lack of reasoning ability of neural networks and graph neural networks to compensate for the logic module's weakness in focusing only on local information. Our model can use both implicit messages from the global perspective and explicit reasoning from the local perspective to make recommendations. In addition, we also designed a more suitable knowledge graph construction method for the model to construct an item-item graph from existing data. Our main contributions are as follows:
\begin{enumerate}
\itemsep=0pt
\item We propose a neural-symbolic recommendation model, which combines graph neural networks with logic reasoning. The model can not only obtain information aggregation gain from the graph but also use propositional logic to reason about users' historical behaviors.
\item We design a new method of constructing graphs for the proposed model, building item-item graphs based on the adjacency principle.
\item We experimented with the proposed model on several real public datasets and compared it with state-of-the-art models. We have also explored different GNN architectures.
\end{enumerate}
\section{Methodology}
\begin{figure}
\includegraphics[width=\textwidth]{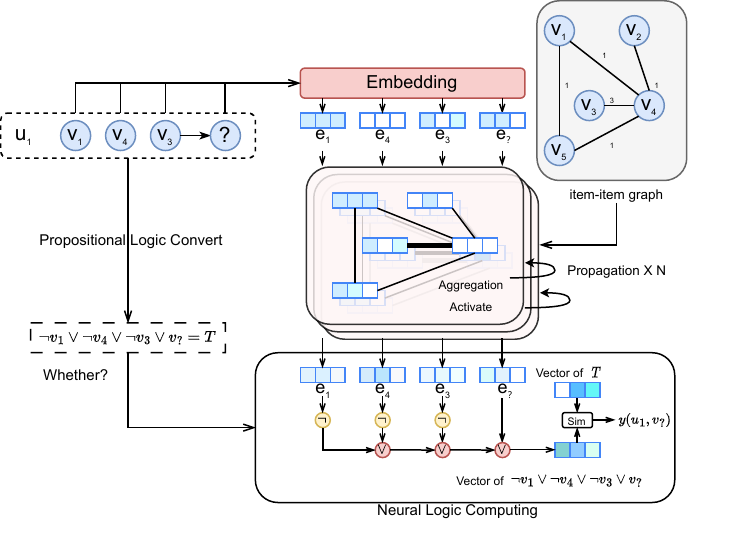}
\caption{GNNLR framework} \label{fig1}
\end{figure}

Fig.~\ref{fig1} illustrates the overall architecture of the proposed model, called GNNLR, which mainly consists of five parts: 1. item-item graph construction; 2. node information fusion; 3. propositional logic convert; 4. neural logic computing; and 5. prediction and training. We will describe these five parts in detail as follows.
\subsection{Item-Item Graph Construction}
\begin{figure}
\includegraphics[width=\textwidth]{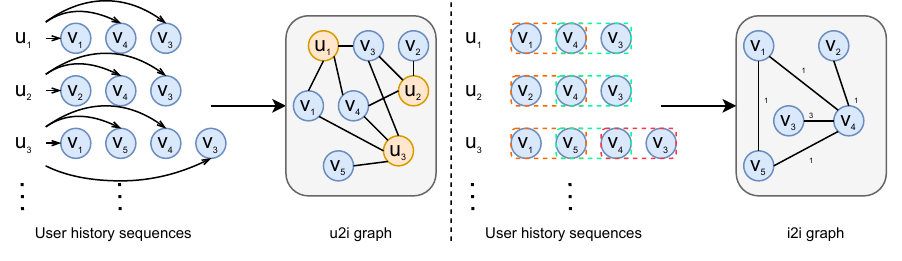}
\caption{Traditional graph construction method (left) and our method (right).} \label{fig2}
\end{figure}

We first describe how the graph required for the model are constructed. We constructs the graph differently from the previous method, as shown in Fig.~\ref{fig2}

Previous method(e.g., NGCF~\cite{wang2019neural}) typically construct a graph based on known user-item interaction relationships (as shown on the left side of Fig.~\ref{fig2}) and use it for subsequent graph neural network calculations. The strength of this approach is that the interaction between the user and the item is retained very directly.However, in a real recommendation scenario, the number of users is often significantly larger than the number of items, and the graph constructed according to the above method will be very sparse and large, because the number of nodes is the number of users added to the number of items, this ultimately affects the performance of the model and causes excessive computational costs. Therefore, we aim to construct a smaller and denser graph that only contains item nodes.

More specifically, for each user's historical interaction sequence, each pair of adjacent items is considered to have an edge. There are two main reasons for considering only adjacent items rather than all items in the same historical interaction sequence: 1. if all items were considered without restrictions, it would be easy for the number of edges to explode. 2. considering only adjacent items can also preserve temporal information. Furthermore, the weight of each edge is the adjacent count of these two items in the history of all user interactions (as shown on the right-hand side of figure ~ref{fig2}). Through the above procedure, we can obtain an undirected weighted homogeneous graph $\mathcal{G}=(\mathcal{V},\mathcal{E})$ with nodes ${{v}_{i}}\in \mathcal{V}$ representing all items and edges $({{v}_{i}},{{v}_{j}})\in \mathcal{E}$ connecting them. We can also obtain a weighted adjacency matrix $A\in {{\mathbb{R}}^{N\times N}}$ and degree matrix ${{D}_{ii}}=\sum\nolimits_{j}{{{A}_{ij}}}$ with $N$ being the number of nodes.
\subsection{Node Information Fusion}
After the item-item graph is constructed, we embed all items as vectors and propagate and aggregate these vectors based on the item-item graph.GNNLR uses a graph convolutional network~\cite{kipf2017} to perform this operation.

Assume that $X$ represents the features of all nodes and $\Theta $ represents all trainable parameters. The formula for obtaining new features in each information aggregation is:
\begin{equation}\label{eq1}
{{X}^{'}}={{\widehat{D}}^{-\frac{1}{2}}}\widehat{A}{{\widehat{D}}^{-\frac{1}{2}}}X\Theta
\end{equation}
where $\widehat{A}=A+I$ indicates that the information of the node itself is kept, $\widehat{D}$ is the degree matrix. For each node ${{x}_{i}}$, its information aggregation formula is:
\begin{equation}\label{eq2}
x_{i}^{'}={{\Theta }^{\top }}\sum\limits_{j\in \mathcal{N}(i)\cup \{i\}}{\frac{{{e}_{j,i}}}{\sqrt{{{\widehat{d}}_{j}}{{\widehat{d}}_{i}}}}{{x}_{j}}}
\end{equation}
Where ${{e}_{j,i}}$ represents the weight of the edge from ${{x}_{j}}$ to ${{x}_{i}}$, and ${{\widehat{d}}_{i}}=1+\sum\nolimits_{j\in \mathcal{N}(i)}{{{e}_{j,i}}}$. In short, each node's features are influenced by its neighbors' features, and the degree of influence is related to the weights of the edges. This aggregation process can be repeated several times, which is beneficial because the information of neighbor nodes' neighbor nodes is also aggregated. In addition, we use the ReLU activation function at the end of each aggregation.
\subsection{Propositional Logic Convert}
After obtaining the aggregated item features, we transform the existing user history behavior into propositional logic expressions to train the model and implement recommendation prediction.

The main symbols of classical propositional logic include $\wedge ,\vee ,\neg ,\to ,\leftrightarrow $, representing 'and', 'or', 'not', 'if...then', and 'equivalent to'. Among them, the operator $\neg $ has the highest calculation priority when there are no parenthesis.

In recommendation tasks, we can naturally convert the historical behavior of each user into propositional logic expressions. For example, if a user has a history of interaction $h=({{v}_{1}},{{v}_{4}},{{v}_{3}})$ and ${{v}_{3}}$ is considered as the target item while ${{v}_{1}},{{v}_{4}}$ is viewed as their historical interaction item, then we can derive the following logic rule:
\begin{equation}\label{eq3}
({{v}_{1}}\to {{v}_{3}})\vee ({{v}_{4}}\to {{v}_{3}})\vee ({{v}_{1}}\wedge {{v}_{4}}\to {{v}_{3}})=T
\end{equation}
Formula 3 represents that a user's preference for ${{v}_{3}}$ may be due to their previous liking of ${{v}_{1}}$, or ${{v}_{4}}$, or because they have liked both ${{v}_{1}}$ and ${{v}_{4}}$ at the same time.

Subsequently, according to the logic implication equivalence $p\to q=\neg p\vee q$, the implied formula can be transformed into a disjunctive normal form consisting of Horn clauses. The above formula can be transformed as:
\begin{equation}\label{eq4}
(\neg {{v}_{1}}\vee {{v}_{3}})\vee (\neg {{v}_{4}}\vee {{v}_{3}})\vee (\neg ({{v}_{1}}\wedge {{v}_{4}})\vee {{v}_{3}})=T
\end{equation}

At this point, the transformed logic expressions still have the problem of computational complexity. When the number of items in the history interaction increases, the length of the expressions and the number of logic variables will explode, especially the number of variables in the higher-order Horn clauses. Based on our calculation, when there are $n$ items in the history interaction, the number of Horn clause terms in this disjunctive normal form is $\sum\limits_{r=1}^{n}{\frac{n!}{r!(n-r)!}={{2}^{n}}-1}$ and its computational complexity is $O({{2}^{n}})$. Therefore, we further simplify those high-order Horn clauses using De Morgan's law. According to De Morgan's law $\neg (p\wedge q)\leftrightarrow \neg p\vee \neg q$, we can further transform Eq.~\ref{eq4} into:
\begin{equation}\label{eq5}
(\neg {{v}_{1}}\vee {{v}_{3}})\vee (\neg {{v}_{4}}\vee {{v}_{3}})\vee (\neg {{v}_{1}}\vee \neg {{v}_{4}}\vee {{v}_{3}})=T
\end{equation}

Finally, according to the propositional logic associative law, we can remove the parentheses and eliminate the duplicate variables. The simplified horn clause is obtained as follows:
\begin{equation}\label{eq6}
\neg {{v}_{1}}\vee \neg {{v}_{4}}\vee {{v}_{3}}=T
\end{equation}

At this point, the computational complexity is reduced to $O(n)$. Similarly, we can transform any length of user historical interaction behavior into simplified propositional logic expression to train the model. When we make a prediction, we only need to construct a propositional logic expression consisting of the user's history interaction and the target item and let the model determine whether the expression is true. For example, based on the above historical interaction, we can let the model determine whether the following logic expression is true:
\begin{equation}\label{eq7}
\neg {{v}_{1}}\vee \neg {{v}_{4}}\vee \neg {{v}_{3}}\vee {{v}_{?}}
\end{equation}
Where ${{v}_{?}}$ is the predicted item, the closer the logic expression becomes to true, the more likely the user is to like item ${{v}_{?}}$.
\subsection{Neural Logic Computing}
In this section, we describe how the model computes the converted logic expressions. We adopt the method from paper NLR~\cite{shi2020neural} and use neural networks to perform logic operations.

Specifically, in GNNLR, the propositional logic expression contains two operators $\neg $ and $\vee $. We train two independent neural network modules $NOT(\cdot )$ and $OR(\cdot ,\cdot )$ to perform their corresponding logic operations, and the neural network modules use a multilayer perceptron structure. If the dimension of the item vector after graph neural network aggregation is d, then for module $NOT(\cdot )$, its input is a d-dimension vector, and its output is also a d-dimension vector representing the logic negation of that vector.
\begin{equation}\label{eq8}
\neg {{e}_{i}}=NOT({{e}_{i}})=W_{2}^{not}\sigma (W_{1}^{not}{{e}_{i}}+b_{1}^{not})+b_{2}^{not}
\end{equation}
For module $OR(\cdot ,\cdot )$, its input is a vector of dimension 2d concatenated from two vectors, and the output is a vector of dimension d representing the result of logic disjunction operation on the two input vectors.
\begin{equation}\label{eq9}
{{e}_{i}}\vee {{e}_{j}}=OR({{e}_{i}},{{e}_{j}})=W_{2}^{or}\sigma (W_{1}^{or}({{e}_{i}}\oplus {{e}_{j}})+b_{1}^{or})+b_{2}^{or}
\end{equation}
Where $W_{2}^{not},W_{1}^{not},W_{2}^{or},W_{1}^{or},b_{2}^{not},b_{1}^{not},b_{2}^{or},b_{1}^{or}$ are all learnable model parameters and $\sigma $ is the activation function. The neural logic computing model will compute propositional logic expression according to its order of operation and ultimately outputs a vector ${{e}_{l}}$ of dimension d representing the expression. As shown at the bottom of Fig.~\ref{fig1}.
\subsection{Prediction and Training}
In this section, we describe how to use the computed vectors of logic expressions for prediction and training. When the GNNLR model is initialised, a benchmark vector $T$ of dimension d is generated. We determine whether a logic expression is true by computing the similarity of vector ${{e}_{l}}$ with vector $T$ and compute the similarity by using the following formula:
\begin{equation}\label{eq10}
Sim({{e}_{l}},T)=sigmoid(\varphi \frac{{{e}_{l}}\cdot T}{\left\| {{e}_{l}} \right\|\times \left\| T \right\|})
\end{equation}
Where $\varphi $ is an optional parameter that can be combined with the Sigmoid function to make the model more flexible when dealing with different datasets, the similarity result ranges from 0 to 1. A result closer to 1 indicates that the logic expression is closer to true, and the target item ${{v}_{?}}$ is more likely to be preferred by users. During training, we adopt a pair-wise learning strategy that for each item ${{v}^{+}}$ liked by a user, we randomly sample an item ${{v}^{-}}$ that has not been interacted with or disliked by the user. We then calculate the loss function based on the following formula:
\begin{equation}\label{eq11}
L=-\sum\limits_{{{v}^{+}}}{\log (sigmoid(p({{v}^{+}})-p({{v}^{-}})))}
\end{equation}
Where $p({{v}^{+}})$ and $p({{v}^{-}})$ are the predicted results of model on item ${{v}^{+}}$ and ${{v}^{-}}$. We also use the logic rule loss ${{\mathcal{L}}_{logic}}$ defined in~\cite{shi2020neural} to constrain the training of logic operator modules, and use regularization terms $\sum\limits_{e\in E}{\left\| e \right\|}_{F}^{2}$ for constraining vector lengths and $\left\| \Theta  \right\|_{F}^{2}$ for constraining parameter lengths. The final loss function is shown in Eq.~\ref{eq12}, where ${{\lambda }_{\mathcal{L}}}$, ${{\lambda }_{l}}$, ${{\lambda }_{\Theta }}$ are the weights of three constraint losses.
\begin{equation}\label{eq12}
L=-\sum\limits_{{{v}^{+}}}{\log (sigmoid(p({{v}^{+}})-p({{v}^{-}})))}+{{\lambda }_{\mathcal{L}}}{{\mathcal{L}}_{logic}}+{{\lambda }_{l}}\sum\limits_{e\in E}{\left\| e \right\|}_{F}^{2}+{{\lambda }_{\Theta }}\left\| \Theta  \right\|_{F}^{2}
\end{equation}
\section{Experiments}
\subsection{Datasets and Evaluation Metrics}
We conducted experiments on five real datasets with different categories and data volumes, including GiftCard, Luxury, Software, Industry from the Amazon review website and MovieLens-100k from the MovieLens website. Tab.~\ref{tab1} shows the specific information of these five datasets, where edge\_num represents the number of item-item edges generated by the method in Section 2.1.
\begin{table}\centering
\caption{General statistical information about the five real-world datasets.}\label{tab1}
\setlength{\tabcolsep}{4mm}{
\begin{tabular}{lccccc}
\hline
Dataset &  User & Item & Interaction & Edge\_num & Density\\
\hline
GiftCard & 459 & 149 & 2972 & 2580 & 4.35\%\\
Software & 1826 & 802 & 12805 & 21408 & 0.874\%\\
Luxury & 3820 & 1582 & 34278 & 7906 & 0.567\%\\
Industry & 11042 & 5335 & 77071 & 75758 & 0.131\%\\
ML100k & 943 & 1682 & 1000000 & 71066 & 6.3\%\\
\hline
\end{tabular}}
\end{table}

Considering that some baseline models are based on sequence algorithms, according to the suggestion in~\cite{chen2021neural}, we adopt a leave-one-out strategy to process and divide the dataset: we sort each user's historical interactions by time and use each user's last two positive interactions as validation set and test set.

We use two metrics, H@K (Hit Rate) and N@K (Normalized Discounted Cumulative Gain), to evaluate the performance of our model. A higher value for H@K indicates that the target item appears more frequently in the top K predicted items, while a higher value for N@K indicates that the target item has a more advanced ranking. We randomly sample 50 negative items for the first three datasets as interference for each correct answer during testing and randomly sample 100 negative items as interference for the last two larger datasets.
\subsection{Comparison Methods}
We will compare the proposed GNNLR with the following baseline models, which cover different recommendation approaches including shallow models, deep models, sequence models, graph neural networks and reasoning models:
\begin{itemize} 
\item \textbf{BMF}~\cite{rendle2009bpr}: A matrix factorization model based on Bayesian personalized ranking, which is a very classic recommendation algorithm.
\item \textbf{NCF}~\cite{he2017neural}: Neural Collaborative Filtering is an improved collaborative filtering algorithm that replaces vector dot products with neural networks and integrates traditional matrix factorization.
\item \textbf{STAMP}~\cite{liu2018stamp}: A popular model that takes into account both short-term attention and long-term user behavior memory. 
\item \textbf{NARM}~\cite{li2017neural}: A powerful sequence recommendation model that combines attention mechanism and gated recurrent networks.
\item \textbf{GRU}~\cite{hidasi2018recurrent}: A powerful sequence recommendation model that applies gated recurrent networks to recommendation algorithms.
\item \textbf{NGCF}~\cite{wang2019neural}: This is a state-of-the-art recommendation model based on GNN, which utilizes graph neural networks for collaborative filtering algorithms. It models user-item interactions as a graph structure and performs information aggregation.
\item \textbf{NLR}~\cite{shi2020neural}: Neural logic reasoning, a neural-symbolic model based on modular propositional logic operation of neural networks. This is a state-of-the-art reasoning-based recommendation framework.
\end{itemize}
\subsection{Parameter Settings}
All models were trained with 200 epochs using the Adam optimizer and a batch-size of 128. The learning rate was 0.001 and early-stopping was conducted according to the performance on the validation set. Both ${{\lambda }_{l}}$ and ${{\lambda }_{\Theta }}$ were set to $1\times {{10}^{-4}}$ and applied to the baseline models equally; ${{\lambda }_{r}}$ was set to $1\times {{10}^{-5}}$. The vector embedding dimension was set to 64 for all baseline models. The maximum history interaction length was set to 5 for sequence-based models. More details can be obtained from the code link provided in the abstract.
\subsection{Recommendation Performance}
Tab.~\ref{tab2} shows the recommendation performance of our model and baseline models on five datasets. The best results are highlighted in bold, while the second-best results are underlined.
\begin{table}\centering
\caption{Performance comparison of all models on five datasets.}\label{tab2}
\setlength{\tabcolsep}{1.4mm}{
\begin{tabular}{lccccccccc}
\hline
Dataset &  Metric & BMF & NCF & SMP & NAM & GRU & NGCF & NLR & Ours\\
\hline
\multirow{4}{*}{GiftCard} & N@10 & 0.3028 & 0.3032 & 0.2926 & 0.3409 & \underline{0.3582} & 0.3169 & 0.3308 & \bfseries{0.3646}\\
 & N@20 & 0.3442 & 0.3381 & 0.3344 & 0.3727 & \underline{0.3988} & 0.3656 & 0.3697 & \bfseries{0.4134}\\
 & H@10 & 0.5772 & 0.5894 & 0.5306 & 0.5918 & \underline{0.5967} & 0.5732 & 0.5813 & \bfseries{0.6057}\\
 & H@20 & 0.7398 & 0.7276 & 0.6939 & 0.7492 & 0.7492 & \underline{0.7520} & 0.7358 & \bfseries{0.7642}\\
\hline
\multirow{4}{*}{Software} & N@10 & 0.2903 & 0.2937 & 0.3524 & \underline{0.3831} & 0.3682 & 0.3339 & 0.3794 & \bfseries{0.4487}\\
 & N@20 & 0.3386 & 0.3424 & 0.3971 & 0.4305 & 0.4125 & 0.3767 & \underline{0.4384} & \bfseries{0.4842}\\
 & H@10 & 0.4582 & 0.4757 & 0.5919 & 0.6554 & \underline{0.6597} & 0.5894 & 0.6433 & \bfseries{0.7513}\\
 & H@20 & 0.6487 & 0.6894 & 0.7681 & \underline{0.8370} & 0.8326 & 0.7587 & 0.8354 & \bfseries{0.8809}\\
\hline
\multirow{4}{*}{Luxury} & N@10 & 0.5075 & 0.4707 & 0.5090 & \underline{0.5205} & 0.5135 & 0.5021 & 0.5189 & \bfseries{0.5541}\\
 & N@20 & 0.5505 & 0.5133 & 0.5459 & \underline{0.5568} & 0.5466 & 0.5306 & 0.5522 & \bfseries{0.5841}\\
 & H@10 & 0.6236 & 0.5951 & 0.7196 & 0.7308 & 0.7340 & 0.6317 & \underline{0.7428} & \bfseries{0.7727}\\
 & H@20 & 0.7969 & 0.7749 & 0.8644 & \underline{0.8740} & 0.8636 & 0.8149 & 0.8684 & \bfseries{0.8907}\\
\hline
\multirow{4}{*}{Industry} & N@10 & 0.2553 & 0.2213 & 0.2383 & 0.2611 & 0.2600 & 0.2526 & \underline{0.2612} & \bfseries{0.3163}\\
 & N@20 & 0.2935 & 0.2492 & 0.2697 & 0.2953 & 0.2944 & 0.2873 & \underline{0.2966} & \bfseries{0.3409}\\
 & H@10 & 0.4138 & 0.3401 & 0.3791 & 0.4147 & 0.4232 & 0.3915 & \underline{0.4253} & \bfseries{0.4934}\\
 & H@20 & 0.5425 & 0.4715 & 0.5037 & 0.5558 & 0.5593 & 0.5293 & \underline{0.5603} & \bfseries{0.6217}\\
\hline
\multirow{4}{*}{ML100k} & N@10 & 0.3578 & 0.3595 & 0.3907 & 0.4084 & 0.4094 & 0.3841 & \underline{0.4151} & \bfseries{0.4239}\\
 & N@20 & 0.4085 & 0.4066 & 0.4303 & 0.4435 & 0.4424 & 0.4259 & \underline{0.4458} & \bfseries{0.4581}\\
 & H@10 & 0.6281 & 0.6338 & 0.6602 & 0.6795 & 0.6752 & 0.6488 & \underline{0.6833} & \bfseries{0.6956}\\
 & H@20 & 0.8184 & 0.8081 & 0.8137 & 0.8210 & 0.8092 & 0.8124 & \underline{0.8215} & \bfseries{0.8296}\\
\hline
\end{tabular}}
\end{table}

The experimental results show that the GNNLR model exhibits the best performance on all four metrics of the five data sets due to its ability to utilize global implicit information from graph neural networks and local explicit reasoning from propositional logic. Sequence-based models (e.g., NARM and GRU) and reasoning-based models (NLR) achieve most of the second-best performance, probably because these models are good at utilizing the temporal information in the data, which we retain during the data processing. In addition, GNNLR outperforms both NGCF, which relies solely on graph neural networks for recommendations, and NLR, which relies solely on neural logic for recommendations, and verifies that our contributions are meaningful from the perspective of ablation experiments. Although NGCF performs not very well, the significant improvement of GNNLR over NLR of recommendation results proves the usefulness of graph neural networks. In conclusion, the experimental results show that our proposed GNNLR model and item-item graph construction method can efficiently combine the advantages of neural and symbolic methods and significantly enhance the recommendation results.
\subsection{Research on Different GNN Model}
For GNNLR, the GNN module is a plug-and-play component. Therefore, we further explored the impact of different GNN architectures on the performance of GNNLR models. In addition to GCN (the GNN architecture used by GNNLR), we selected five other different GNN models for testing:
\begin{itemize} 
\item \textbf{GAT}~\cite{wang2019kgat}: The Graph Attention Network.
\item \textbf{Light-GNN}~\cite{he2020lightgcn}: The Light Graph Convolution (LGC) operator.
\item \textbf{ChebGCN}~\cite{defferrard2016convolutional}: The Chebyshev spectral Graph Convolutional operator. 
\item \textbf{GCN2Conv}~\cite{chen2020simple}: The Craph Convolutional operator with initial residual connections and identity mapping.
\item \textbf{FAConv}~\cite{bo2021beyond}: The Frequency Adaptive Graph Convolution operator.
\end{itemize}
\begin{table}\centering
\caption{Comparison of GNNLR with different GNN architectures.}\label{tab3}
\setlength{\tabcolsep}{2mm}{
\begin{tabular}{lccccccc}
\hline
Dataset &  Metric & GCN & GAT & LGNN & ChebGC & GC2N & FAC\\
\hline
\multirow{4}{*}{GiftCard} & N@10 & \bfseries{0.3646} & 0.3352 & 0.3284 & \underline{0.3593} & 0.3547 & 0.3437\\
 & N@20 & \bfseries{0.4134} & 0.3846 & 0.3751 & 0.4023 & \underline{0.4098} & 0.3822\\
 & H@10 & \bfseries{0.6057} & 0.5913 & 0.5688 & 0.5991 & 0.5913 & \underline{0.5994}\\
 & H@20 & 0.7642 & 0.7558 & 0.7517 & \bfseries{0.7683} & \underline{0.7682} & 0.7539\\
\hline
\multirow{4}{*}{Software} & N@10 & \bfseries{0.4487} & 0.4318 & 0.4325 & \underline{0.4342} & 0.4291 & 0.3915\\
 & N@20 & \bfseries{0.4842} & 0.4708 & 0.4746 & \underline{0.4776} & 0.4693 & 0.4345\\
 & H@10 & \bfseries{0.7513} & \underline{0.7379} & 0.7208 & 0.7204 & 0.7236 & 0.6824\\
 & H@20 & 0.8809 & \underline{0.8873} & 0.8805 & \bfseries{0.8879} & 0.8826 & 0.8526\\
\hline
\multirow{4}{*}{Luxury} & N@10 & \bfseries{0.5541} & 0.5486 & \underline{0.5511} & 0.5405 & 0.4908 & 0.4692\\
 & N@20 & \bfseries{0.5841} & \underline{0.5782} & 0.5714 & 0.5729 & 0.5219 & 0.5104\\
 & H@10 & \underline{0.7727} & 0.7723 & \bfseries{0.7814} & 0.7611 & 0.7152 & 0.6793\\
 & H@20 & \underline{0.8907} & 0.8788 & \bfseries{0.8947} & 0.8879 & 0.8573 & 0.8410\\
\hline
\multirow{4}{*}{Industry} & N@10 & \bfseries{0.3163} & 0.2881 & 0.3003 & 0.2871 & \underline{0.3009} & 0.2635\\
 & N@20 & \bfseries{0.3409} & 0.3242 & \underline{0.3349} & 0.3206 & 0.3335 & 0.2970\\
 & H@10 & \bfseries{0.4934} & \underline{0.4832} & 0.4818 & 0.4747 & 0.4770 & 0.4266\\
 & H@20 & \underline{0.6217} & \bfseries{0.6260} & 0.6195 & 0.6070 & 0.6062 & 0.5595\\
\hline
\multirow{4}{*}{ML100k} & N@10 & \bfseries{0.4239} & 0.3939 & 0.4127 & \underline{0.4178} & 0.3840 & 0.3864\\
 & N@20 & \bfseries{0.4581} & 0.4359 & 0.4524 & \underline{0.4566} & 0.4215 & 0.4285\\
 & H@10 & \bfseries{0.6956} & 0.6602 & 0.6763 & \underline{0.6849} & 0.6624 & 0.6517\\
 & H@20 & 0.8296 & 0.8253 & \bfseries{0.8328} & \underline{0.8382} & 0.8103 & 0.8189\\
\hline
\end{tabular}}
\end{table}
The experimental results using different GNN modules are shown in Tab.~\ref{tab3}. The traditional GCN architecture achieves the best results in most metrics, Light-GNN, ChebGCN and GAT also showed the best performance on some metrics. This indicates that different GNN architectures have their own advantages when facing different types of data or recommendation metrics. As for the worse performance of the GCN2Conv and FAConv models, we thought it might be due to the complex structure that takes more time to converge. We used a uniform number of epochs in our experiments, and more epochs may further improve the performance of these two GNN architectures.
\section{Conclusion}
In this work, we propose a Neural-Symbol recommendation model that combines the advantages of graph neural networks and logic reasoning, named GNNLR. GNNLR uses both global implicit information from graphs and local explicit reasoning from propositional logic for recommendation prediction. We also design a method for constructing item-item graphs for GNNLR better to integrate graph neural networks with propositional logic reasoning. We conduct extensive experiments on five real-world datasets and explore the effects of different graph neural network architectures on GNNLR performance. Extensive experiments verified the effectiveness of the GNNLR model; and showed that different graph neural network architectures have their advantages when facing datasets with different characteristics.

In future work, we will explore and construct more graphs with different perspectives and combine them to enable graph neural networks to further extract rich global implicit information from multiple perspectives. Meanwhile, we will incorporate user information in the logic reasoning module and utilize first-order logic to enhance its flexibility and scalability.

\subsubsection{Acknowledgements} This work was supported by the National Natural Science Foundation of China (No. 61906066), Natural Science Foundation of Zhejiang Province (No. LQ18F020002), Zhejiang Provincial Education Department Scientific Research Project(No. Y202044192), Postgraduate Research and Innovation Project of Huzhou University (No. 2022KYCX43).

%
%
%
\bibliographystyle{splncs04}
\bibliography{springer}
%






\end{document}